\documentclass[letterpaper, 10 pt, conference]{ieeeconf}
\pdfoutput=1
\IEEEoverridecommandlockouts
\usepackage{times}
\usepackage{amsmath}
\usepackage{amssymb}
\usepackage{algorithm}
\usepackage{graphics}
\usepackage{epsfig}
\usepackage[tight]{subfigure}
\usepackage{tabularx}
\usepackage{booktabs}
\usepackage{multirow}
\usepackage{xspace}
\usepackage{flushend}
\usepackage{color}
\usepackage[textsize=scriptsize,disable]{todonotes}
\newcommand{\ourmodel}[0]{Ours\xspace}
\newcommand{\ourmodelmid}[0]{Ours-Mid\xspace}
\makeatletter
\let\NAT@parse\undefined
\makeatother
\usepackage[numbers,sort&compress]{natbib}
\bibpunct{[}{]}{,}{n}{}{;}

\title{\LARGE \bf Satellite Image-based Localization via Learned Embeddings}
\author {Dong-Ki Kim \hspace{20pt} Matthew R. Walter \thanks{Dong-Ki Kim is with the Massachusetts Institute of Technology, Cambridge, MA USA, {\tt\small dkkim93@mit.edu}} \thanks{Matthew R.\ Walter is with the Toyota Technological Institute at Chicago, Chicago, IL USA, {\tt\small mwalter@ttic.edu}}}
\usepackage[hidelinks,
pdfauthor={Dong-Ki Kim and Matthew R. Walter},
pdftitle={Satellite Image-based Localization via Learned Embeddings},]{hyperref}
\makeatletter
\hypersetup{pdftitle={\@title},pdfauthor={Dong-Ki Kim and Matthew R. Walter}}
\makeatother

\begin{document}
\maketitle
\begin{abstract}
    We propose a vision-based method that localizes a ground vehicle
    using publicly available satellite imagery as the only prior
    knowledge of the environment. Our approach takes as input a
    sequence of ground-level images acquired by the vehicle as it navigates,
    and outputs an estimate of the vehicle's pose relative to a georeferenced
    satellite image. We overcome the significant viewpoint
    and appearance variations between the images through a neural
    multi-view model that learns
    location-discriminative embeddings in which ground-level images are
    matched with their corresponding satellite view of the scene. We
    use this learned function as an observation model in a filtering
    framework to maintain a distribution over the vehicle's pose. We
    evaluate our method on different benchmark datasets and
    demonstrate its ability localize ground-level images in environments novel relative
    to training, despite the challenges of significant viewpoint and appearance
    variations. The video highlight is available at \textcolor{blue}{\url{https://youtu.be/58K1-0WpGNs}}.
\end{abstract}

\section{Introduction} \label{sec:intro}
Accurate estimation of a vehicle's position and orientation is
integral to autonomous operation across a broad range of applications
including intelligent transportation, exploration, and
surveillance. Currently, many vehicles employ Global Positioning
System (GPS) receivers to estimate their absolute, georeferenced
pose. However, most commercial GPS systems suffer from limited
precision, are sensitive to multipath effects (e.g., in the
so-called ``urban canyons'' formed by tall buildings), which can
introduce significant biases that are difficult to detect, or may not
be available (e.g., due to jamming). Visual
place recognition seeks to overcome these limitations by identifying a
camera's (coarse) pose in an a priori known environment (typically
in combination with map-based localization, which uses visual
recognition for loop-closure). Visual place recognition is
challenging due to the appearance variations that result from changes
in perspective, scene content (e.g., parked cars that are no longer
present), illumination (e.g., due to the time of day), weather, and
seasons. A number of techniques have been proposed that
make significant progress towards overcoming these
challenges~\citep{churchill12, milford12, johns13, sunderhauf13, naseer14, mcmanus14,
 sunderhauf15, sunderhauf15a}. However, most methods perform localization relative
to a database of geotagged ground images, which requires that the
environment be mapped a priori.
\begin{figure}[!t]
    \centering
    \includegraphics[width=0.8\linewidth]{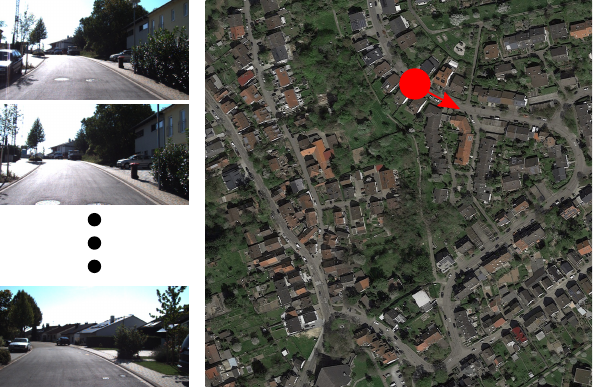}
    \caption{Our model estimates a vehicle's pose
    on a georeferenced satellite image (right) given input of a sequence of
    ground-level images (left).}
    \label{fig:motivation}
\end{figure}

Satellite imagery provides an alternative source of information that
can be employed as a reference for vehicle localization~\citep{jacobs07, bansal11, lin13, viswanathan14, viswanathan16}.
High resolution, georeferenced, satellite images that densely cover the world are becoming increasingly accessible and well-maintained, as exemplified by Google Maps. The goal is then to
perform visual localization using satellite images as the only prior
map of the environment. However, registering ground-level images to their corresponding location in a
satellite image of the environment is challenging. The difference in
their viewpoints means that content visible in one type of image is
often not present in the other. For example, whereas ground-level
images include building fa{\c c}ades and tree trunks, satellite images
include roofs and the tops of trees. Additionally, the dynamic nature of
the scene means that objects will differ between views. In
street scenes, for example, the same parked and stopped cars as well as
pedestrians that make up a large fraction of the objects visible at ground-level are
not present in satellite views. Meanwhile, satellite imagery may
have been acquired at different times of the day and at different
times of the year, resulting in appearance variations between
ground-level and satellite images due to illumination, weather, and seasons.

In this paper, we describe a framework that employs multi-view
learning to perform accurate vision-based localization using satellite
imagery as the only prior knowledge of the environment. Our system
takes as input a sequence of ground-level images acquired as a vehicle navigates and returns an estimate of
its location and orientation in a georeferenced satellite image (Fig.~\ref{fig:motivation}).
Rather than matching the query images to a prior map of geotagged ground-level images, as is
typically done for visual place recognition, we describe a neural
multi-view Siamese network that learns to associate novel ground
images with their corresponding position and orientation in a
satellite image of the scene. We investigate the use of both
high-level features, which reduce viewpoint invariance, and mid-level
features, which have been shown to exhibit greater invariance to
appearance variations~\cite{sunderhauf15a} as part of our network architecture. As we show, we can train
this learned model on ground-satellite pairs from one environment and employ the model in a different environment, without the need for ground-level
images for fine-tuning. The framework uses outputs of this learned matching function as observations
in a particle filter that maintains a distribution over the vehicle's
pose. In this way, our approach exploits the availability of satellite
images to enable visual localization in a manner that is robust to
disparate viewpoints and appearance variations, without the need for
a prior map of ground-level images. We evaluate our method on the
KITTI~\cite{kitti} and St.\ Lucia~\cite{warren10} datasets, and demonstrate the ability to transfer our learned,
hierarchical multi-view model to novel environments and thereby
localize the vehicle, despite the challenges of severe viewpoint
variations and appearance changes.

\section{Related Work}
The problem of estimating the location of a query ground image is typically framed as
one of finding the best match against a database (i.e., map) of
geotagged images. In general, existing approaches broadly fall into
one of two classes depending on the nature of the query and database images.

\subsection{Single-View Localization}
Single-view approaches assume access to reference databases that consist of
geotagged images of the target environment acquired from vantage
points similar to that of the query image (i.e.,
other ground-level images). These databases may come in the form of
collections of Internet-based geotagged images, such as those
available via photo sharing websites~\citep{zhang06, hays08, crandall09}
or Google Street View~\citep{zamir10, majdik13}, or
maintained in maps of the environment (e.g., previously recorded using
GPS information or generated via SLAM)~\citep{schindler07, cummins08, chen11a, churchill12, badino12, milford12,
johns13, sunderhauf13,  mcmanus14, sunderhauf15, sunderhauf15a}.
The primary challenges to visual place recognition arise due to variations in viewpoint,
variations in appearance that result from changes in environment
structure, illumination, and seasons, as well as to perceptual
aliasing. Much of the early work attempts to mitigate some of these
challenges by using hand-crafted features that exhibit some robustness
to transformations in scale and rotation, as well as to slight
variations in illumination (e.g., SIFT~\cite{sift} and
SURF~\cite{surf}), or a combination of visual and textual (i.e., image
tags) features~\citep{crandall09}. Place recognition then
follows as image retrieval, i.e., image-to-image matching-based search
against the database~\citep{wolf05, li06, filliat07, schindler07}.

These techniques have proven effective at identifying the location of
query images over impressively large areas~\citep{hays08}.
However, their reliance upon available reference images limits their use to regions with sufficient coverage and their
accuracy depends on the spatial density of this coverage. Further,
methods that use hand-crafted interest point-based features tend to
fail when faced with significant viewpoint and appearance variations,
such as due to large illumination changes (e.g., matching a query
image taken at night to a database image taken during the day) and
seasonal changes (e.g., matching a query image with snow to one taken
during summer). Recent attention in visual place recognition~\citep{valgren07, glover10, badino12, neubert13, mcmanus14, maddern14, lowry14, sunderhauf15, sunderhauf15a} has focused on designing algorithms
that exhibit improved invariance to the challenges of viewpoint and
appearance variations. Motivated by their state-of-the-art performance on object detection and recognition tasks~\cite{krizhevsky12}, a solution that has proven
successful is to use deep convolutional networks to learn suitable
feature representations. \citet{sunderhauf15a} present a thorough
analysis of the robustness of different AlexNet~\cite{krizhevsky12}
features to appearance and viewpoint variations. Based on these
insights, \citet{sunderhauf15} describe a framework that first
detects candidate landmarks in an image and then employs mid-level
features from a convolutional neural network (AlexNet) to perform place recognition despite
significant changes in appearance and viewpoint. We also employ CNNs
as a means of learning mid- and high-level features that exhibit
improved robustness to viewpoint and appearance variations.
Meanwhile, an alternative to single-view matching is to consider image
sequences when performing recognition~\citep{koch10, milford12, johns13, sunderhauf13, naseer14}, whereby imposing joint consistency reduces the likelihood of false matches and improves robustness to appearance variations.

\subsection{Cross-View Localization}
Cross-view methods identify the location of
a query image by matching against a database of images taken from
disparate viewpoints. As in our case, this typically involves
localizing ground-level images using a database of georeferenced
satellite images~\citep{jacobs07, bansal11, lin13, viswanathan14, viswanathan14, chu15, lin15, workman15, viswanathan16}. \citet{bansal11} describe a method
that localizes street view images relative to a collection of geotagged oblique
aerial and satellite images. Their method uses the combination of
satellite and aerial images to extract building fa{\c c}ades and their
locations. Localization then follows by matching these fa{\c c}ades against
those in the query ground image. Meanwhile, \citet{lin13} leverage the
availability of land use attributes and propose a cross-view learning
approach that learns the correspondence between hand-crafted
interest-point features from ground-level images, overhead images, and
land cover data. \citet{chu15} use a database of ground-level images
paired with their corresponding satellite views to learn a dictionary
of color, edge, and neural features that they use for retrieval at
test time.  Unlike our  method, which requires only satellite
images of the target environment, both approaches assume access to
geotagged ground-level images from the test environment that are used
for training and matching.

\citet{viswanathan14} describe an algorithm that warps 360$^\circ$
view ground images to obtain a projected top-down view that they then
match to a grid of satellite locations using traditional hand-crafted
features. As with our framework, they use the inferred poses as
observations in a particle filter. The technique assumes that the
ground-level and satellite images were acquired at similar times, and
is thus not robust to the appearance variations that arise as a result
of seasonal changes. \citet{viswanathan16} overcome this limitation by
incorporating ground vs.\ non-ground pixel classifications derived
from available LIDAR scans, which improves robustness to seasonal
changes. As with their previous work, they also employ a Bayesian
filter to maintain an estimate of the vehicle's pose. In contrast,
our method uses only an odometer and a non-panoramic, forward-facing
camera with a much narrower field-of-view. Rather than interest point
features, we learn to separate location-discriminative feature
representations for ground-level and satellite views. These features
include an encoding of the scene's semantic properties, serving a
similar role to their ground labels.

Similar to our work is that of \citet{lin15}, who describe a Siamese
Network architecture that uses two CNNs to transform ground-level and
aerial images into a common feature space. Localizing a query ground
image then corresponds to finding the closest georeferenced aerial
image in this space. They train their network on a database of
ground-aerial pairs and demonstrate the ability to localize test
images from environments not encountered during training. Unlike our
method, they match against 45$^\circ$ aerial images, which share more
content with ground-level images (e.g., building fa{\c c}ades) than do
satellite views. Additionally, whereas their Siamese network
extends AlexNet~\cite{krizhevsky12} by using only the second-to-last
connected layer (fc7) as the high-level feature, our network adapts VGG-16~\cite{Simonyan15}
with modifications that consider the use of both
mid-level  and high-level features to improve robustness to changes in appearance.

\section{Proposed Approach} \label{sec:approach}
\begin{figure}[t]
    \centering
    \includegraphics[height=0.48\linewidth]{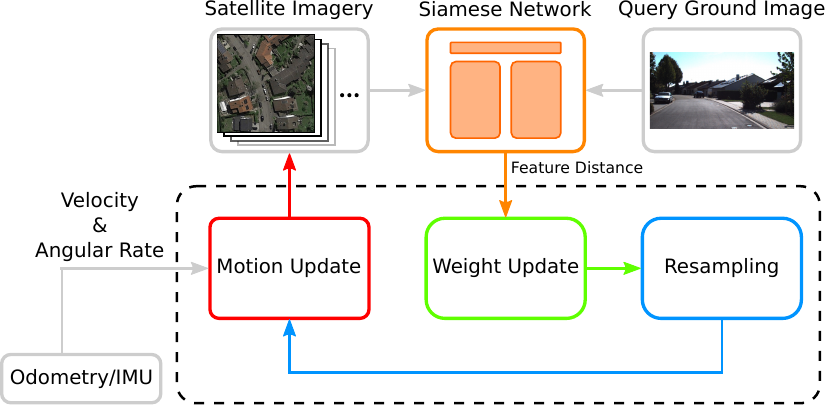}
    \caption{Our method takes as input a stream of ground-level images
      and maintains a distribution over the vehicle's pose by
      comparing these images to a database of satellite images.}
    \label{fig:pipeline}
\end{figure}

Our visual localization framework (Fig.~\ref{fig:pipeline}) takes as
input a stream of ground-level images $\mathcal{I}_g^t = \{\ldots,
I_{t-2}, I_{t-1}, I_t\}$ from a camera mounted to the
vehicle and proprioceptive measurements of the vehicle's motion (e.g.,
from an odometer and IMU), and outputs a distribution over the
vehicle's pose $x_t$ relative to a database of georeferenced satellite
images $\mathcal{I}_s$, which constitutes the only prior knowledge of the
environment. The method consists of a Siamese network that learns
feature embeddings suitable to matching ground-level imagery with
their corresponding satellite view. These matches then serve as noisy
observations of the vehicle's position and orientation that are then
incorporated into a particle filter to maintain a distribution over
its pose as it navigates. Next, we describe these components in detail.

\subsection{Siamese Network} \label{sec:siamese}
\begin{figure}[!t]
    \centering
    \includegraphics[height=0.48\linewidth]{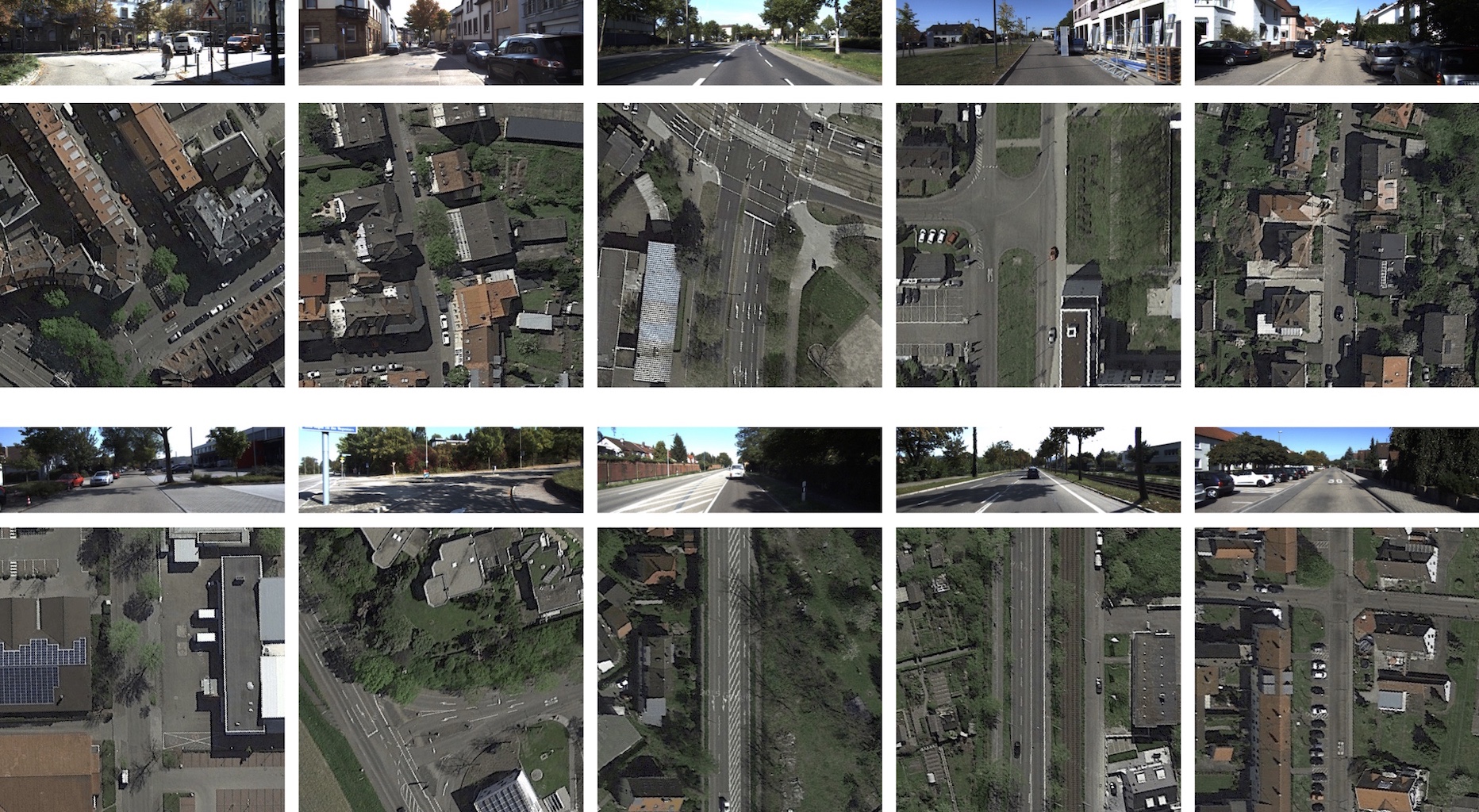}
    \caption{Examples of ground-level images (top rows) paired with their corresponding satellite views (bottom rows). Best viewed in electronic form.}
    \label{fig:ground-truth-pairs}
\end{figure}

Our method matches a query ground-level image to its corresponding
satellite view using a Siamese
network~\cite{Chopra05}, which has proven effective at different
multi-view learning tasks~\cite{taigman14}. Siamese networks consist of two
convolutional neural networks (CNNs), one for each of the two disparate
views. The two networks operate in parallel, performing nonlinear
transformation of their respective input (images) into a common
embedding space, constituting a learned feature representations for
each view. A query is then matched by finding the
nearest cross-view in this embedding space.
\begin{figure*}
    \centering
    \includegraphics[width=0.8\textwidth]{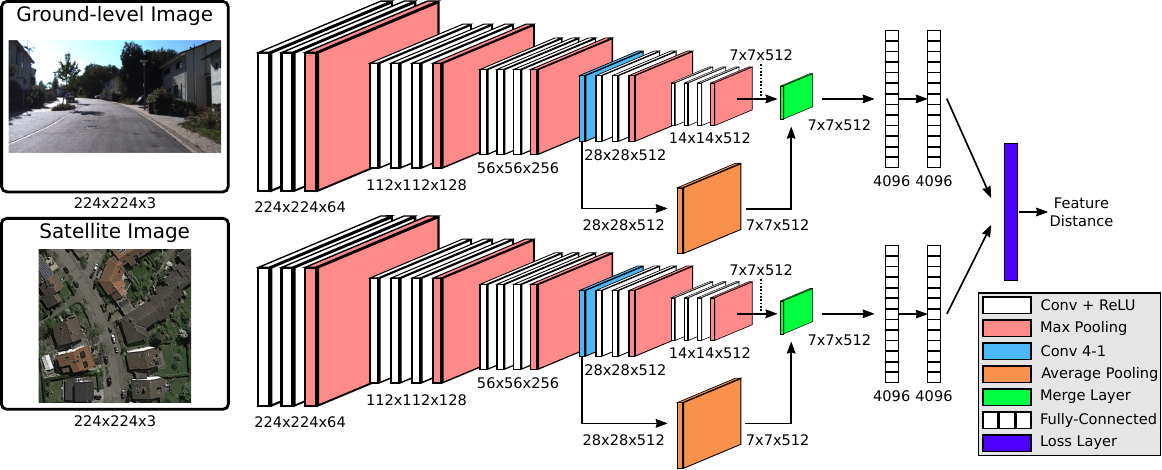}
    \caption{A visualization of our network architecture that consists
    of two independent CNNs that take as input ground-level and
    satellite images. Each CNN is an adaptation of VGG-16 CNNs in which mid-level
   conv4-1 features are downsampled and combined with the output of
   the last max-pooling layer as the high-level features via
   summation. The resulting outputs are then used as a measure of
   distance between ground-level and satellite views.} \label{fig:siamese}
\end{figure*}

Our architecture extends a Siamese network to learn
location-discriminative feature representations that differentiate
between matched ground-satellite pairs and non-matched
ground-satellite pairs. The network (Fig.~\ref{fig:siamese}) takes as
input a pair of ground-level and satellite images that are fed into
their respective convolutional neural networks, which output a
$d$-dimensional feature embedding ($d=4096$) of each image. As we
describe below, the network is trained to discriminate between
positive and negative view-pairs using a loss that encourages positive
matches to be nearby in the learned embedding space, and features to
be distant for non-matching views. Figure~\ref{fig:ground-truth-pairs} presents
examples of ground-level images and their matching satellite views,
demonstrating the challenging viewpoint and appearance variations. At test time, a query ground-level
image is projected into this common space and paired with the
satellite view whose embedding is closest in terms of Euclidean distance.

Our network consists of identical CNN architectures for each of the
two views that take the form of VGG-16 networks~\cite{Simonyan15},
with modifications to improve robustness to variations in
appearance. The first modification removes the softmax layer and the last fully connected layer from the original VGG-16 network and adds a batch
normalization layer to arrive at the $4096$-dimensional high-level
feature representation. The second
modification that we consider incorporates additional mid-level
features into the learned representation. Mid-level features have been shown to
exhibit greater invariance to changes in appearance that result from
differences in illumination, weather, and seasons, while high-level
output features provide robustness to viewpoint
variations~\cite{sunderhauf15a}. Specifically, we use the output from
the conv4-1 layer as the mid-level features and the output from the
last max-pooling layer as the high-level feature representation. We
note that conv4-1 features of VGG-16 are qualitatively similar to
conv3 of AlexNet~\cite{krizhevsky12}, which \citet{sunderhauf15a} use
for place recognition, in terms of their input
stimuli~\cite{zeiler14}. Additionally, high-level features encode semantic information, which is particularly useful for categorizing scenes, while mid-level
features tend to be better suited to discriminating between instances
of particular scenes.  In this way, the two feature representations are
complementary. We downsample the conv4-1 mid-level
features to match the size of the output from the last max-pooling
layer using average-pooling, and combine the two via
summation.\footnote{We explored other methods for combining the two representations including concatenation on a validation set, but found summation to yield the best performance.}
We evaluate the effect of combining these features in Section~\ref{sec:results}.

We train our network so as to learn nonlinear transformations of input
pairs that bring embeddings of matching ground-level and satellite
images closer together, while driving the embeddings associated with
non-matching pairs farther apart. To that end, we use a contrastive
loss~\cite{hadsell06}
\begin{equation} \label{eqn:loss}
    \!\!L(I_g,I_s,\ell)=\ell d (I_g,I_s)^2 + (1\!-\ell) \max(m-d (I_g,I_s),0)^2
\end{equation}
to define the loss between a pair of ground-level $I_g$ and satellite
$I_s$ images, where $\ell$ is a binary-valued correspondence variable that indicates
whether the pair match ($\ell=1$), $d (I_g,I_s)$ is the Euclidean distance
between the image feature embeddings, and $m>0$ is a
margin for non-matched pairs.

In the case of matching pairs of ground-level and satellite images
$(\ell=1)$, the loss encourages the network to learn feature
representations that are nearby in the common embedding space. In the
case of non-matching pairs ($\ell=0$), the contrastive loss penalizes
the network for transforming the images into features that are
separated by a distance less than the margin $m$. In this way,
training will tend to pull matching ground-level and satellite image
pairs closer together, while keeping non-matching images at least a
radius $m$ apart, which provides a form of regularization against
trivial solutions~\cite{hadsell06}.

The ground-level and satellite CNN networks can share model parameters
or be trained separately. We chose to train the networks with
separate parameters for each CNN, but note that previous efforts found
little difference with the addition of parameter sharing in similar
Siamese networks~\cite{lin15}.

\subsection{Particle Filter} \label{sec:particleFilter}
The learned distance function modeled by the convolutional neural
network provides a good measure of the similarity between ground-level
imagery and the database of satellite views. As a result of perceptual
aliasing, however, it is possible that the match that minimizes the
distance between the learned features does not correspond to the
correct location. In order to mitigate the effect of this noise, we
maintain a distribution over the vehicle's pose as it navigates the
environment
\begin{equation}
    p(x_t \vert u^t, z^t),
\end{equation}
where $x_t$ is the pose of the vehicle and $u^t = \{u_0, u_1, \ldots, u_t\}$ denotes the history of proprioceptive measurements (i.e., forward velocity and angular
rate). The term $z^t = \{z_0, z_1, \ldots, z_t\}$ corresponds to the history of image-based observations, each consisting of the distance $d (I_g,I_s)$ in the common embedding space between the transformed ground-level image $I_t$ and the satellite image $I_s \in \mathcal{I}_s$ with a position and orientation specified by
$x_t$.\footnote{In the case of forward-facing ground-level cameras, the location associated with the center of the satellite image is in front of the vehicle.}

The posterior over the vehicle's pose tends to be
multimodal. Consequently, we represent the distribution using a set of
weighted particles
\begin{equation}
    \mathcal{P}_t = \left\{P_t^{(1)}, P_t^{(2)}, \ldots, P_t^{(n)} \right\},
\end{equation}
where each particle $P_t^{(i)} = \{x_t^{(i)}, w_t^{(i)}\}$ includes a
sampled vehicle pose $x_t^{(i)}$ and the particle's importance weight
$w_t^{(i)}$.

We maintain the posterior distribution as new odometry and
ground-level images arrive using a particle
filter. Figure~\ref{fig:pipeline} provides a visual overview of this
process. Given the posterior distribution $p(x_{t-1}\vert u^{t-1}, z^{t-1})$ over the vehicle's pose at time $t-1$, we first compute the prior distribution over $x_t$ by
sampling from the motion model prior $p(x_t \vert x_{t-1}, u^t, z^{t-1})$, which we model as Gaussian.

After proposing a new set of particles, we update their weights to
according to the ratio between the target (posterior) and proposal
(prior) distributions
\begin{equation}
    \tilde{w}_t^{(i)} = p(z_t \vert x_t, u^t, z^{t-1}) \cdot w_{t-1}^{(i)},
\end{equation}
where $\tilde{w}_t^{(i)}$ denotes the unnormalized weight at time
$t$. We use the output of the Siamese network to define the
measurement likelihood as an exponential distribution
\begin{equation}
    p(z_t \vert x_t, u^t, z^{t-1}) = \alpha e^{-\alpha d (I_t,I_s)},
\end{equation}
where $d(I_t, I_s)$ is the Euclidean distance between the CNN
embeddings of the current ground-level image $I_t$ and the satellite
image $I_s$ corresponding to pose $x_t$, and $\alpha$ is a tuned
parameter.

After having calculated and normalized the new importance weights, we
periodically perform resampling in order to discourage particle
depletion based on the effective number of particles
\begin{equation}
    N_{\text{eff}} = \frac{1}{\sum\limits_{i=1}^{N} {w_t^{(i)}}^2}
\end{equation}
Specifically, we use systematic resampling~\cite{carpenter99} when the effective number of particles $N_{\text{eff}}$ drops below a threshold.

\section{Results} \label{sec:results}
We evaluate our model through a series of experiments conducted on two
benchmark, publicly available visual localization datasets. We
consider two versions of our architecture: \ourmodelmid uses both mid- and high-level features,
while \ourmodel uses only high-level features.
The experiments analyze the ability of our framework to generalize to different test environments and to mitigate appearance variations that result from changes in semantic scene content and illumination.

\subsection{Experimental Setup}
Our evaluation involves training our network on pairs of ground-level
and satellite images from a portion of the KITTI~\cite{kitti}
dataset, and testing our method on a different region from KITTI and the St.~Lucia dataset~\cite{warren10}.

\subsubsection{Baselines}
We compare the performance of our model against two baselines that
consider both hand-crafted and learned feature representations. The
first baseline (SIFT) extracts SIFT features~\cite{sift} from each
ground-level and satellite image and computes the distance between a
query ground-level image and the extracted satellite image as the
average distance associated with the best-matching features. The
second baseline (AlexNet-Places) employs a Siamese network comprised of two AlexNet
deep convolutional networks~\cite{krizhevsky12} trained on the Places
dataset~\cite{zhou14}. We use the $4096$ dimensional output of the fc7
layer as the embedding when computing the distance used in the
measurement update step of the particle filter.

\subsubsection{Training Details}
Our training data is drawn from the KITTI dataset, collected from a
moving vehicle in Karlsruhe, Germany during the daytime in the months
of September
and October. The dataset consists of sequences that span
variations of city, residential, and road scenes. Of these, we
randomly sample $18$ sequences from the city, residential, and road categories as the training set,
and $5$ from the city and road categories as the validation set, resulting in $14.8$k and $1.3$k ground-level images, respectively.

For each ground-level image in the training and validation sets, we sample a $270 \times 400$ ($53\text{\,m} \times 78\text{\,m}$) satellite image at a position
$5.0\text{\,m}$ in front of the camera with the long-axis oriented in the direction of the camera.
We also include a randomly sampled set of non-matching satellite images. We define a pair of ground-level and satellite images $(I_g, I_s)$ to be a match $(\ell = 1)$
if their distance is within $4$\,m and their orientation within $30$
degrees. We consider non-matches $(\ell = 0)$ as those that are more than
$80$\,m apart.\footnote{These parameters were tuned on the validation set.}
The resulting training set then consists of $538$k pairs ($53$k positive pairs and $485$k negative pairs).
Figure~\ref{fig:ground-truth-pairs} presents samples drawn from the set of positive pairs, which demonstrate the challenge of matching these disparate views.

We trained our models from scratch\footnote{We also tried fine-tuning from models pre-trained on ImageNet and Places and found the results to be comparable.} using the contrastive loss (Eqn.~\ref{eqn:loss})
with a margin of $m=80$ (tuned on the validation set) using Adam~\cite{adam} on an Nvidia Titan X.
Meanwhile, we used the validation set to tune hyper-parameters including the early stopping criterion, as well as to
explore different variations of our network architecture, including alternative methods for combining mid- and high-level features and the use of max-pooling instead of average-pooling.

\subsubsection{Particle Filter Implementation}
In each experiment, we used $N=5000$
particles representing samples of the vehicle's position and
orientation. We assumed that the vehicle's initial pose was unknown
(i.e., the kidnapped robot setting),
and biased the initialization of each filter such that
samples were more likely to be drawn on roadways. Particles were
resampled using $N_{\text{eff}} = 0.8 N$ (tuned on a validation
set). We determine a filter to have converged when the standard
deviation of the estimates is less than $10$\,m.

\subsection{Experimental Results}
We evaluate two versions of our method against the different baselines with regards to
the effects of appearance variations due to changes in
viewpoint, location, scene content, and illumination both
between training and test as well as between the reference satellite
and ground-level imagery. We analyze the performance  in terms of precision-recall as well as
retrieval. We then investigate the accuracy with which the
particle filter is able to localize the vehicle as it navigates.

\subsubsection{KITTI Experiment}
We evaluate our method on KITTI using two residential sequences (KITTI-Test-A and
KITTI-Test-B) as test sets. We
note that there is no environment overlap between these sequences and
those used for training or validation.
The georeferenced satellite maps for
KITTI-Test-A is $0.80\,\text{km} \times 1.10\,\text{km}$ and
$0.53\,\text{km} \times 0.70\,\text{km}$ for KITTI-Test-B.
\begin{figure}[!t]
    \centering
    \subfigure[Precision-Recall]{\includegraphics[width=\linewidth]{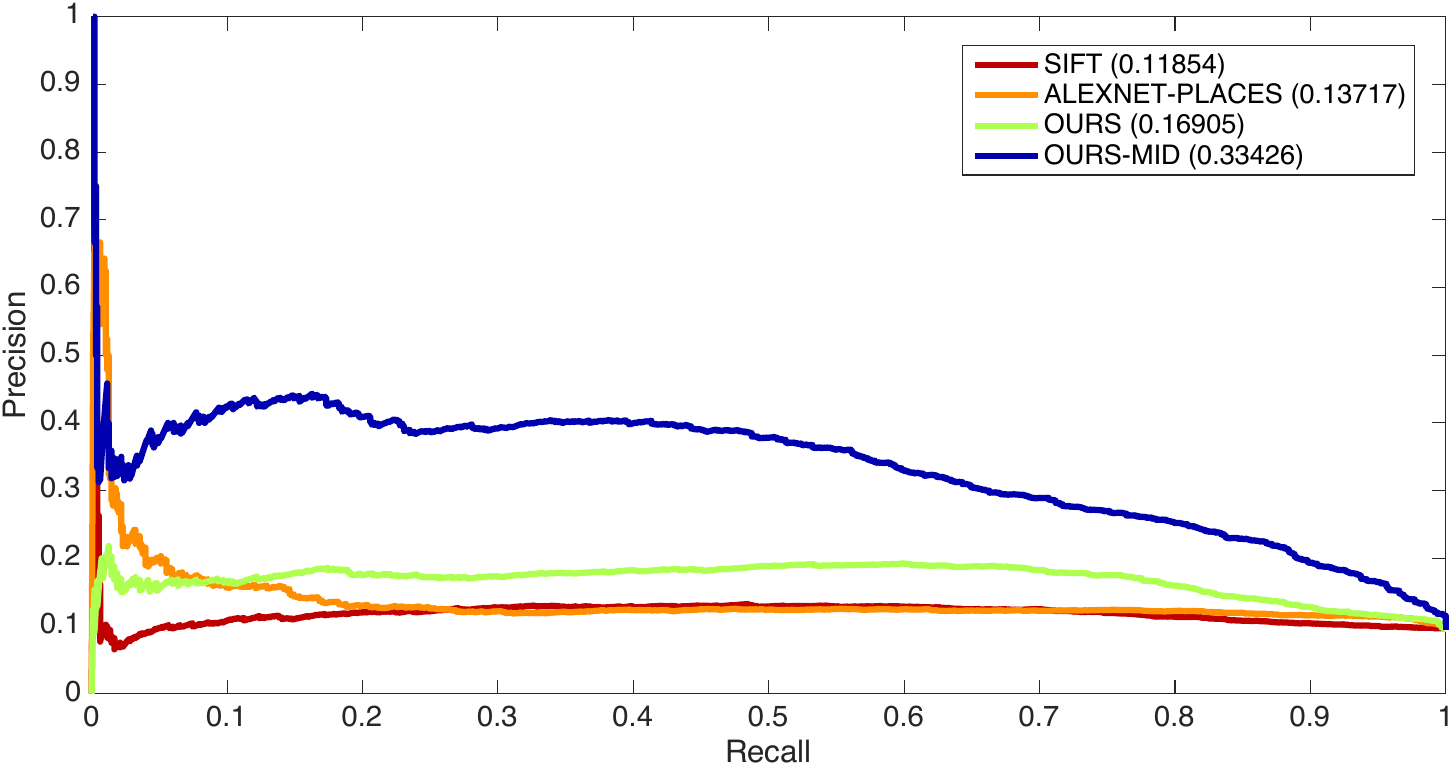}\label{fig:kitti-test-pr}}\\
    \subfigure[Retrieval Accuracy]{\includegraphics[width=\linewidth]{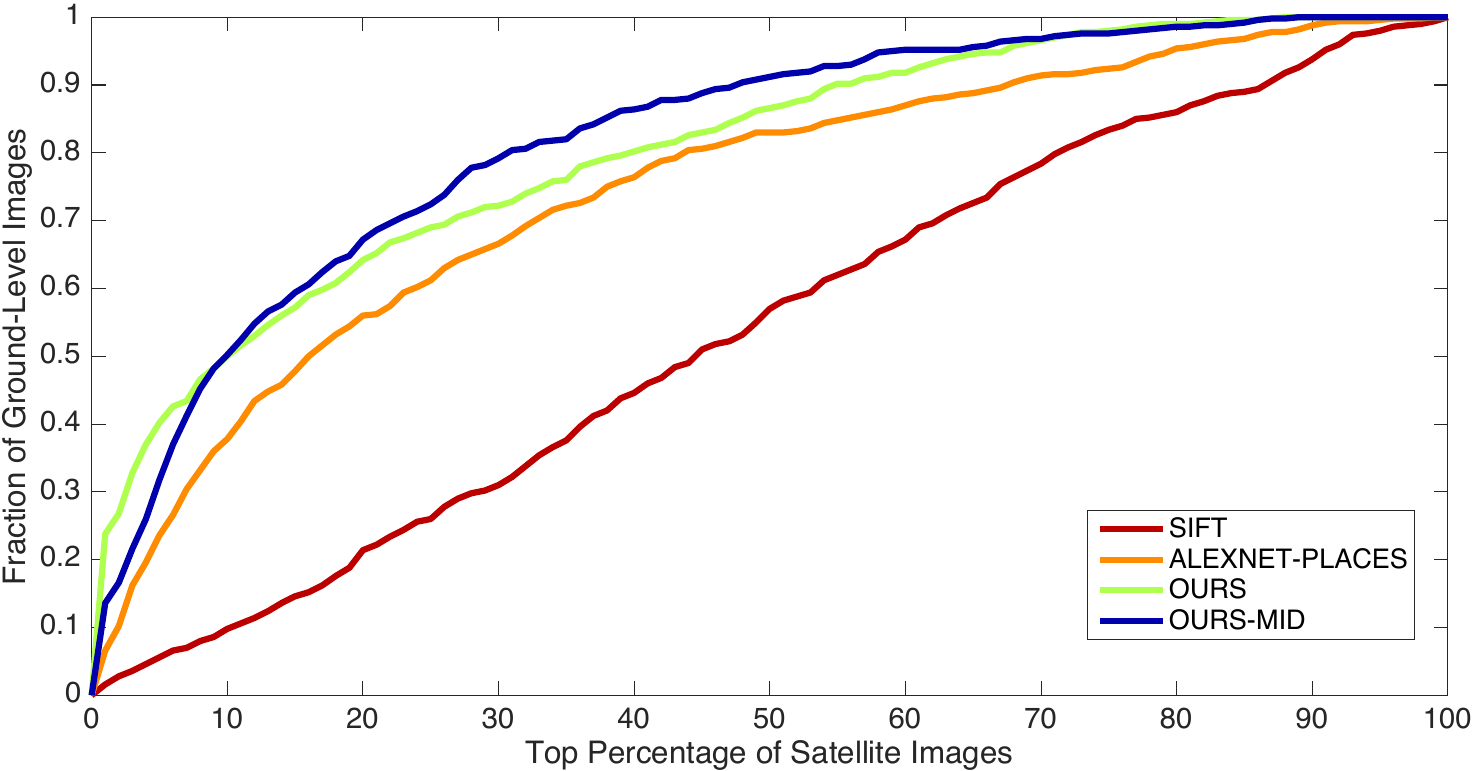}\label{fig:kitti-test-topk}}
    \caption{Results on the KITTI test set including
      \subref{fig:kitti-test-pr} precision-recall curves with the
      average precision values in parentheses and
      \subref{fig:kitti-test-topk} retrieval accuracy.}
\end{figure}

Figure~\ref{fig:kitti-test-pr} compares the precision-recall curve of
our models against the two baseline for the two
KITTI test sequences. The plot additionally includes the average
precision for each model. The plot shows that \ourmodelmid is
the most effective at matching ground-level images with their
corresponding satellite views. The use of a contrastive loss as a
means of encouraging our network to bring positive pairs closer
together facilitates accurate matching. Comparing against the
performance of \ourmodel demonstrates that the incorporation of mid-level features helps to mitigate the
effects of appearance variations. Meanwhile, the AlexNet-Places
baseline that uses the output of AlexNet trained on Places as the
learned feature embeddings outperforms the SIFT baseline that
relies upon hand-designed SIFT features to identify correspondences.

As another measure of the discriminative ability of our network
architecture, we consider the frequency with which the correct satellite
view is in the top-$X\%$ matches for a given ground-level image
according to the feature embedding
distance. Figure~\ref{fig:kitti-test-topk} plots the fraction of query
ground-level images for which the corresponding satellite view is
found in the top-$X\%$ of the satellite images. In the case of both \ourmodelmid and \ourmodel, the correct satellite view is in the top
$10\%$ for more than $50\%$ of the ground-level images. When we
increase the size of the candidate set, our method yields a slight
increase in performance (around $4\%$). The AlexNet-Places
baseline performs slightly worse, while all three significantly
outperform the SIFT baseline, which is essentially equivalent to
chance.
\begin{figure*}[!t]%
    \centering
    \includegraphics[height=2.9in]{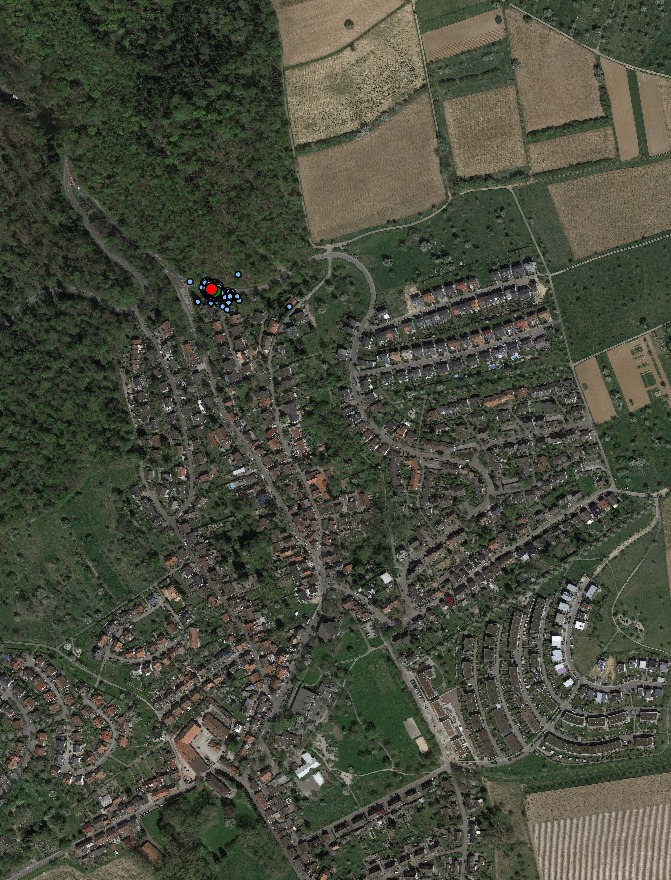}\hfil
    \includegraphics[height=2.9in]{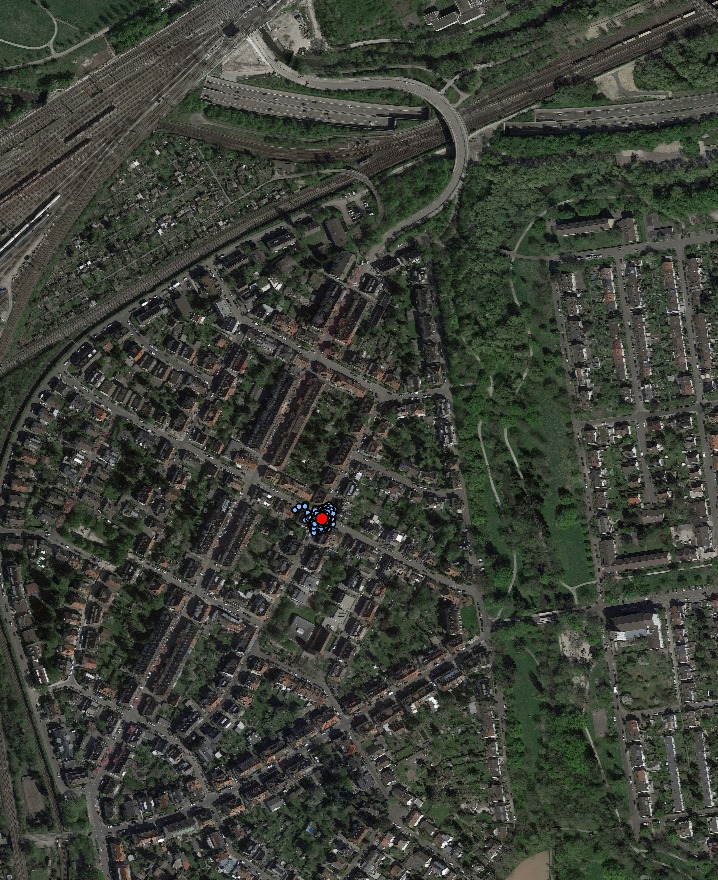}\hfil
    \includegraphics[height=2.9in]{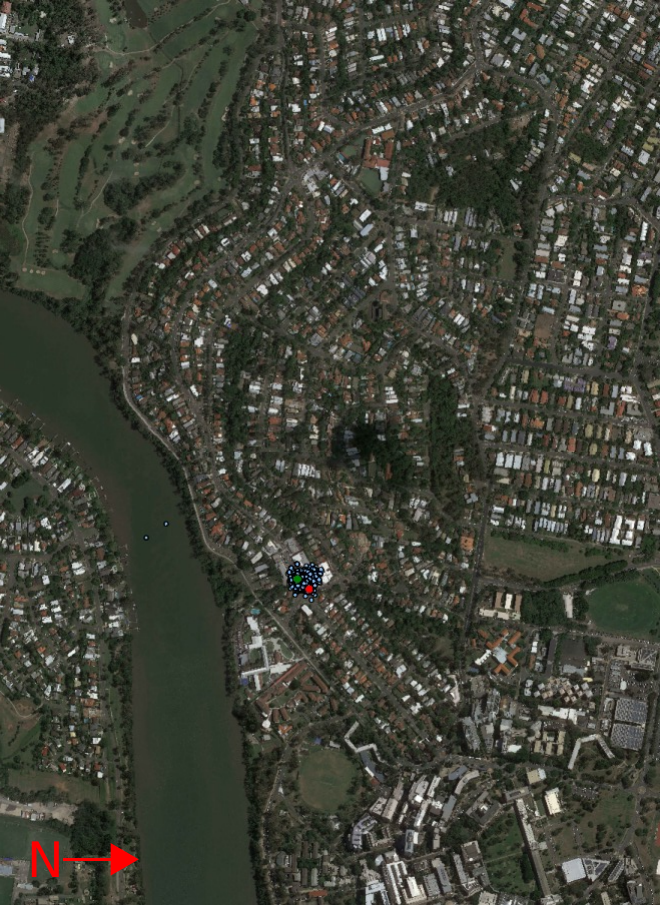}
    \vspace{-6pt}
    \caption{Particle filter localization for KITTI-Test-A
      (left), KITTI-Test-B (center), and St-Lucia
      (right) with blue circles denoting the position associated with
      each particle. The green and red circles indicate the mean position
      estimate and ground-truth, respectively. Note that St-Lucia is
      rotated.} \label{fig:kitti-st-lucia-test-particles}%
      \vspace{-11pt}
\end{figure*}
\begin{table}[!t]
    \centering
	\caption{\label{tab:main} Final mean position error and standard
      deviation (in meters)}
    \setlength{\tabcolsep}{5pt}
    \begin{tabular}{lccc}
      \toprule
        & KITTI-Test-A & KITTI-Test-B & St-Lucia\\
      \midrule
        SIFT & $656.70$ ($244.29$) & $243.08$ ($79.63$) & $554.35$ ($17.26$)\\
        AlexNet-Places & $177.82$ \hphantom{$0$}($25.68$) & $59.95$ ($38.99$) & $77.15$ ($52.61$) \\
        Ours & $8.41$ ($5.56$) & $7.93$ ($2.14$) & $\mathbf{26.38}$ ($5.63$) \\
        Ours-Mid & $\mathbf{7.69}$ ($5.14$) & $\mathbf{4.65}$ ($2.77$) & $35.81$ ($7.54$) \\
      \bottomrule
      \end{tabular}
\end{table}

Next, we evaluate our method's ability to estimate the vehicle's pose
as it navigates the environment by using the distance between the
learned feature representations as observations in the particle filter.
In Table~\ref{tab:main}, we report the quantitative results for each
of the two KITTI test environments. Note that the final position mean
error and standard deviation is measured at the last ground-level
image sequence using all of the particles.
Figure~\ref{fig:kitti-st-lucia-test-particles} depicts the converged particle filter estimate of the vehicle's position using \ourmodelmid compared to the ground-truth position.
For both KITTI-Test-A and KITTI-Test-B, the filter that uses \ourmodelmid has smaller average position errors,
compared to the one that uses \ourmodel. Using \ourmodelmid, the filter converged at $55.62$\,s and $62.25$\,s for {KITTI-Test-A} and {KITTI-Test-B}, respectively, and using \ourmodel, the filter
converged at $64.67$\,s and $70.15$\,s for {KITTI-Test-A} and {KITTI-Test-B}, respectively.
Meanwhile, the SIFT baseline failed to converge with large average errors for both KITTI-Test-A and KITTI-Test-B.
The AlexNet-Places baseline faired better, but the filter also did not converge for either of the test sets.
The results support the argument that our method's ability to learn mid- and high-level feature representations with a loss
that brings ground-level image embeddings closer to their corresponding satellite representation results in measurements
that are able to distinguish between correct and incorrect matches.

\subsubsection{St.\ Lucia Experiment}
Next, we consider the St.\ Lucia dataset~\cite{warren10} as a means of
evaluating the method's ability to generalize the model trained on
KITTI to new environments with differing semantic content. The
dataset was collected during August from a car driving through a
suburb of Brisbane, Australia at different times during a single day
and over different days during a two week period. We use the dataset
collected on August $21$ at $12$:$10$ as the validation set and the
dataset collected on August $19$ at $14$:$10$ as the test set. The two
exhibit slight variations in viewpoint due to differences in the
vehicle's route. More pronounced appearance variations result from
illumination changes and non-stationary objects in the environment
(e.g., cars and pedestrians). Note that the dataset does not
include the vehicle's velocity or angular rate. As done
elsewhere~\cite{glover10} we simulate visual odometry by interpolating
the GPS differential, which is prone to significant noise and is
thereby representative of the quality of data that visual odometry
would yield. The georeferenced satellite map is $1.8$\,km $\times 1.2$\,km.

Table~\ref{tab:main} compares the performance on the St.\ Lucia
dataset. Figure~\ref{fig:kitti-st-lucia-test-particles} (right)
presents the converged particle filter estimate maintained using our
method (\ourmodelmid), along with the ground-truth vehicle location
during the run. Despite the viewpoint and appearance variations
relative to the training set and between the ground-level and
satellite views, both the filters using \ourmodelmid and \ourmodel
maintain a converged estimate of the vehicle's pose. The filter
associated with \ourmodelmid resulted in a larger final position error
than that associated with \ourmodel. Using \ourmodelmid, the filter
converged $17\%$ faster than that using \ourmodel. Both the SIFT and
AlexNet-Places baselines failed to converge.

\subsubsection{Computational Efficiency}
As with other approaches to visual localization~\cite{mcmanus14a, sunderhauf15}, the current implementation of our framework does not process images in real-time.
Computing the CNN feature representations for a pair of ground-level and satellite images requires approximately \mbox{$55$\,ms} on an Nvidia Titan X, though the two network components can be decoupled,
thereby requiring that each ground-level image be embedded only once, which reduces overall computational cost.
For the experiments presented in this paper, we cropped and processed the satellite image associated with each particle at every update step.
This process can be made significantly more efficient by pre-processing satellite images corresponding to a discrete set of poses,
in which case the primary cost is that of computing the ground-level image embeddings. Empirical results demonstrate that this results in a negligible decrease in accuracy.

\section{Conclusion} \label{sec:conclusion}
We described a method that is able to localize a ground vehicle by
exploiting the availability of satellite imagery as the only prior map
of the environment. Underlying our framework is a multi-view neural
network that learns to match ground-level images with their
corresponding satellite view. The architecture enables our method to
learn feature representations that help to mitigate the
challenge of severe viewpoint variation and that improve robustness to
appearance variations resulting from changes in illumination and
scene content.  Distances in this common embedding space then serve as
observations in a particle filter that maintain an estimate of the
vehicle's pose. We evaluate our model
on benchmark visual localization datasets and demonstrate the ability
to transfer our learned multi-view model to novel environments. Future work includes adaptations to the
mult-view model that tolerate more severe appearance variations (e.g.,
due to seasonal changes).

\section{Acknowledgements} \label{sec:ack}

We thank the NVIDIA corporation for the donation of Titan X GPUs used in this research.

\bibliographystyle{IEEEtranN}
\setlength{\bibsep}{0pt plus -0.3ex} {\small \bibliography{references}}

\end{document}